\documentclass[letterpaper]{article} % DO NOT CHANGE THIS
\usepackage{aaai25}  % DO NOT CHANGE THIS
\usepackage{times}  % DO NOT CHANGE THIS
\usepackage{helvet}  % DO NOT CHANGE THIS
\usepackage{courier}  % DO NOT CHANGE THIS
\usepackage[hyphens]{url}  % DO NOT CHANGE THIS
\usepackage{graphicx} % DO NOT CHANGE THIS
\usepackage{natbib}  % DO NOT CHANGE THIS AND DO NOT ADD ANY OPTIONS TO IT
\usepackage{caption} % DO NOT CHANGE THIS AND DO NOT ADD ANY OPTIONS TO IT
\usepackage{algorithm}
\usepackage{algorithmic}
\usepackage{amsmath}
\usepackage{subcaption}
\usepackage{multirow}
\usepackage{booktabs}
\usepackage{pifont}
\usepackage[table]{xcolor}

\newcommand{\cmark}{\textcolor{green!80!black}{\ding{51}}}
\newcommand{\xmark}{\textcolor{red}{\ding{55}}}
\usepackage{newfloat}
\usepackage{listings}
\DeclareCaptionStyle{ruled}{labelfont=normalfont,labelsep=colon,strut=off} % DO NOT CHANGE THIS
\floatstyle{ruled}
\newfloat{listing}{tb}{lst}{}
\floatname{listing}{Listing}
\NewDocumentCommand\emojifloodsynth{}{\includegraphics[scale=0.25]{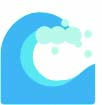}}
\title{\emojifloodsynth MultiFloodSynth: Multi-Annotated Flood Synthetic Dataset Generation}

\author{
    YoonJe Kang\textsuperscript{\rm 1}\equalcontrib,
    Yonghoon Jung\textsuperscript{\rm 1}\equalcontrib,
    Wonseop Shin\textsuperscript{\rm 1}\equalcontrib,
    Bumsoo Kim\textsuperscript{\rm 1}\equalcontrib,
    Sanghyun Seo\textsuperscript{\rm 1}\thanks{Corresponding author}\\
}
\affiliations{
    \textsuperscript{\rm 1}Chung-Ang University, Republic of Korea\\
    \{bluejay100, dydgns2017, wonseop218, bumsookim, sanghyun$^\dagger$\}@cau.ac.kr\\
}

\begin{document}

\maketitle

\thispagestyle{plain}

\begin{abstract}
In this paper, we present synthetic data generation framework for flood hazard detection system. For high fidelity and quality, we characterize several real-world properties into virtual world and simulate the flood situation by controlling them. For the sake of efficiency, recent generative models in image-to-3D and urban city synthesis are leveraged to easily composite flood environments so that we avoid data bias due to the hand-crafted manner. Based on our framework, we build the flood synthetic dataset with 5 levels, dubbed \textit{MultiFloodSynth} which contains rich annotation types like normal map, segmentation, 3D bounding box for a variety of downstream task. In experiments, our dataset demonstrate the enhanced performance of flood hazard detection with on-par realism compared with real dataset.
\end{abstract}

\section{Introduction}

Deep learning algorithm requires high quality, diverse and large training dataset. However, gathering good dataset is labor-intensive and requires substantial cost, especially on hyper-scale situations like wildfire recognition  \cite{hong2024wildfire}, pine wilt disease detection  \cite{refjung2024harnessing} and so on. To mitigate this issue, a common solution today is to generate synthetic data and use them as training or reference dataset \cite{kim2024early}.  Recently, generative models  \cite{hamza2024ali, islam2024diffusemix, khullar2023synthetic} or real-time engine  \cite{delussu2024synthetic} are leveraged for more plausible and efficient generation.

However, there is a still concern that editability is insufficient (that is few control parameter) to composite the final scene.  Existing dataset has ambiguous label and one or two types of annotation. Furthermore, from the fact that synthetic dataset should reflect the real-world data, some domains have a significant difficulty due to the absence of high quality real data for reference, definition of label and inconsistent annotation problem. One of them is flood hazard situation which makes it difficult to collect the dataset. It stems from their specific situation where flood accidents frequently paralyze the digital system and physically collapse the surveillance device. Despite of this reason, existing works had explored to make real flood dataset with hand-crafted labeling \cite{wan2024automatic, floodwu2024identification, floodgao2024measuring}. They inevitably confront label-inconsistent problem like 2D bounding box. In addition, they only consider one or two types of annotation as ground truth, hindering their applicability to various computer vision tasks.

In this paper, we present a novel framework that utilizes a 3D engine to generate urban flood synthetic dataset, dubbed \textit{MultiFloodSynth}. For fidelity, we faithfully attribute the flood hazard situation as several properties and components (e.g., layout   \cite{shang2024urbanworld}, lighting, flood-level   \cite{chaudhary2020water, wan2024automatic}, 3D object, camera view) for scene composition by exploring urban flood situation. Moreover, considering various computer vision tasks, our system includes a variety of annotation types such as normal map, instance/semantic/fine-grained segmentation map, camera 3D pos, 2D/3D bounding box, etc. Thanks to such editable attributes, our \textit{MultiFloodSynth} improved the performance of object-localized flood level detection, while alleviating large dataset requirements for model training.

\section{Related Works}
\subsection{Synthetic Dataset Generation}
\label{sec:rel_work_syntehtic}

Recently, generating non-real dataset, synthetic dataset is common technique in a variety of field which requires hyper-scale  \cite{shang2024urbanworld, ref25Greff2022KubricAS, refzhang2024cityx, refxie2024citydreamer, refwu2024unique3d, refschieber2024indoor, shang2024urbanworld, wang2024high, hao2024synthetic, zhu2024odgen, valvano2024controllable}, impossible scenarios     \cite{refjung2024harnessing,ref25Greff2022KubricAS, refhummel2019leveraging, mittal2023orbit, kokosza2024scintilla, amador2024cyclogenesis}, and so on. Synthetic dataset generation resolve such issues by constructing scene in virtual world and alleviate vexing manual process with auto labeling. Diverging from conventional way to generate synthetic data, it has been explored to reflect the characteristics of real-world object to enhance the high fidelity and appropriateness   \cite{refrichter2022enhancing, lee2024learning, ebadi2022psp}. Since these strategy enhance the robustness and realism, it is crucial to consider these attributes for faithful synthesized data.

\begin{figure*}[t]
    \centerline{\includegraphics[width=1\linewidth]{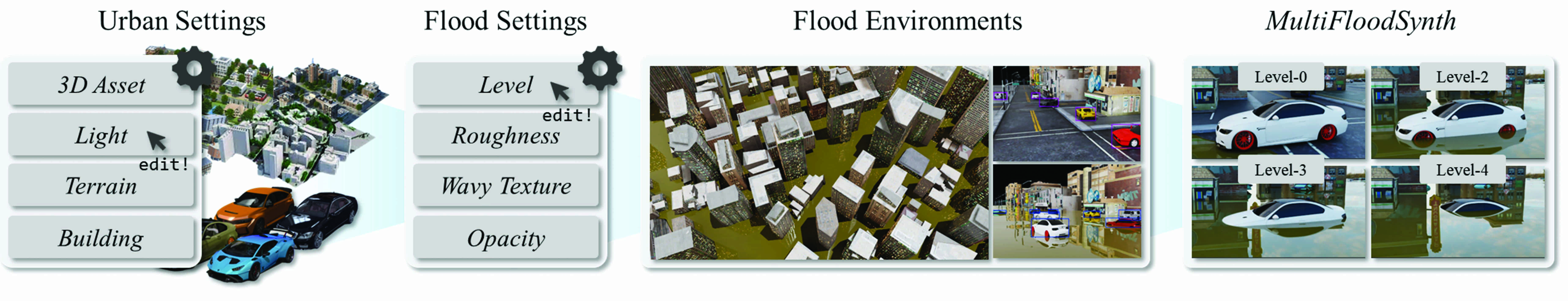}}
    \caption{Overview of virtual flood scene composition and synthetic dataset generation pipeline.}
    \label{fig:overview}
\end{figure*}

\begin{figure}[t]
    \centerline{\includegraphics[width=0.9\linewidth]{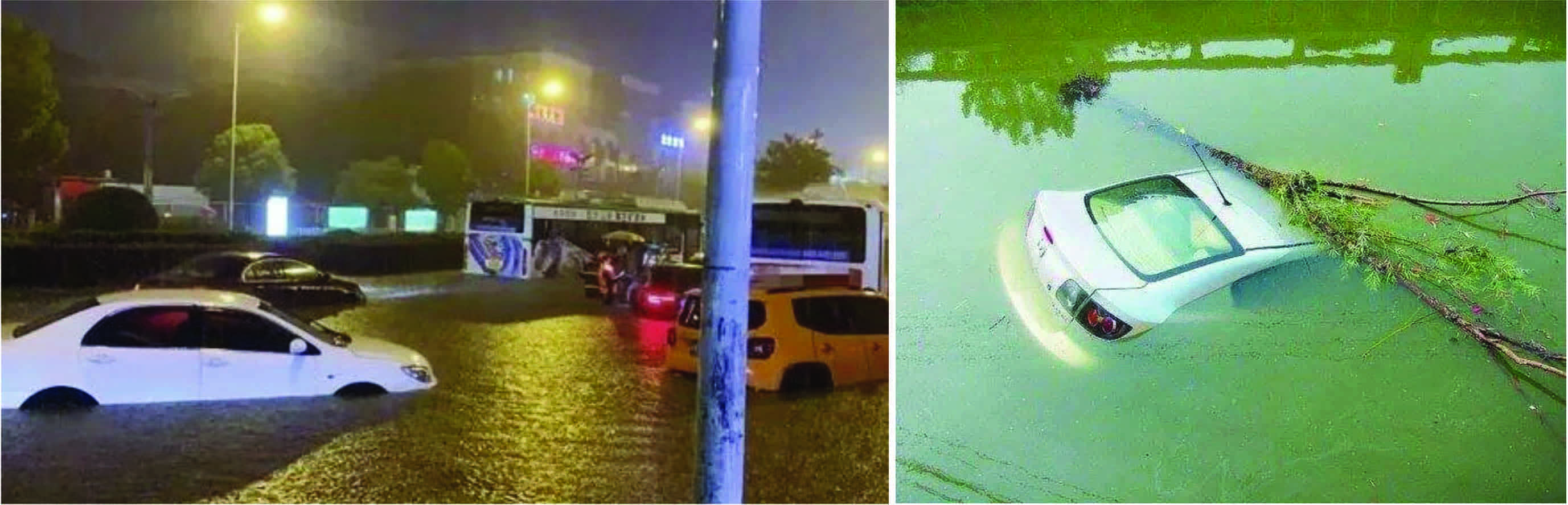}}
    \caption{Sample of \textit{real} flood data   \cite{wan2024automatic}.}
    \label{fig:real_flood}
\end{figure}

\begin{figure}[t]
    \centerline{\includegraphics[width=0.9\linewidth]{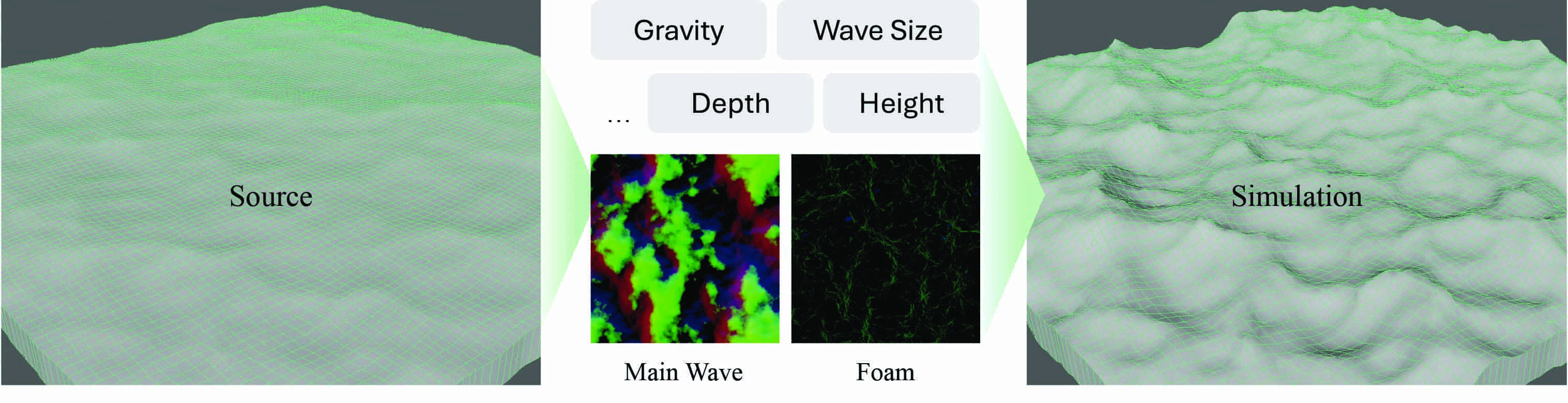}}
    \caption{Components of flood simulation and results.}
    \label{fig:flood_simulation}
\end{figure}

\subsection{Flood Hazard Detection} \label{sec:rel_work_flood}

Detecting flood situation can be considered as object detection using the level of flood of object. A multitude of studies on classifying and detecting objects based on deep learning algorithms has been continuously conducted to address abovementioned requirement  \cite{floodlo2021deep, floodpally2022application, floodkaranjit2023floodimg, floodzhong2024detection, floodwu2024identification}. However, most studies face challenges such as a lack of data sharing, ethical concerns, absence of abundance and limited labels. To do that, in this paper, we intent to address these issues in following section.

\section{Proposed Method} \label{sec:method}

Main objective is to synthesize urban-scale flood hazard situation in virtual scene and enhance the performance of flood-level detection system with our \textit{MultiFloodSynth} by alleviating a problem of dataset collection. To commence, we describe the real-world flood hazard situation with some considerations and how our \textit{MultiFloodSynth} promise the fidelity. Overall pipeline is shown in Fig. \ref{fig:overview}.

\subsection{Challenges of Real-World Flood Hazard Scenarios}
Existing real-world dataset \cite{floodgao2024measuring} was obtained on vehicle-based flood detection. This dataset includes label information divided into five levels based on the percentage of a vehicle submerged in water \cite{wan2024automatic}. Fig. \ref{fig:real_flood} shows some samples of the data included in the real-world dataset. However, they have inconsistent problem with incoherent bounding box by human-hand. Meanwhile, to simulate the flood situation, it should be considered to appropriately composite several components including camera view, lighting condition and flood-level (\textit{i.e.}, flood height).

\begin{table}[t] 
    \centering
    \begin{tabular}{l | l | l }
    \toprule
    \textbf{Parameter} & \textbf{Attribute} & \textbf{Type} \\
    \midrule
    \multicolumn{3}{c}{\textit{Urban Settings}} \\
    \midrule
    Position & Transform & Constant \\
    Lighting & Light Intensity & Constant \\
    Background & Texture & Image \\
    Layout & - & Image \\
    \midrule
    \multicolumn{3}{c}{\textit{Flood Settings}} \\
    \midrule
    Level (Scale) & Transform & Constant \\
    Roughness & - & Constant \\
    Wavy Texture & - & Image \\
    Opacity & Material & Constant \\
    Specular & Material & Constant \\
    Main Wave & - & Image \\
    Wave Foam & - & Image \\
    \bottomrule
    \end{tabular}
    
    \caption{Control parameters to composite the virtual flood hazard scene. $-$ denotes no corresponding attribute.}
    \label{tab:parameters}
\end{table}

\begin{table*}[t]
    \centering
    \begin{tabular}{l | c c c c c c c}
    \toprule
    & \cite{floodzhong2024detection} & \cite{floodgao2024measuring} &  \cite{floodwu2024identification} &  \cite{wan2024automatic} &  \textit{MultiFloodSynth} \\
    Data type & \textit{real} & \textit{real} & \textit{real} & \textit{real} &\textit{synthetic}\\
    \midrule
    
    Flood-Level & 4 & \xmark & 3 & 5 & 5\\
    Availability & \xmark & \xmark & \xmark & \cmark & \cmark$^\dagger$ \\ 
    \midrule
    Frames & 1,177 & 45,199 & 19,062 & 2,000 & 70,117 \\
    Labeling  &Manual {\scriptsize(Inconsistent)}&Manual {\scriptsize(Inconsistent)}&Manual {\scriptsize(Inconsistent)}& Manual {\scriptsize(Inconsistent)} & Auto {\scriptsize(Consistent)}\\
    \midrule
    Normal Map & \xmark & \xmark & \xmark & \xmark & \cmark \\
    Camera Pos & \xmark & \xmark & \xmark & \xmark & \cmark \\
    3D BBox & \xmark & \xmark & \xmark & \xmark & \cmark \\
    2D BBox & \cmark & \cmark & \xmark & \cmark & \cmark \\
    Depth Map & \xmark & \xmark & \xmark & \xmark & \cmark \\
    Semantic Seg & \xmark & \cmark & \xmark & \xmark & \cmark \\
    Instance Seg & \xmark & \xmark & \xmark & \xmark & \cmark \\
    Fine-grained Seg & \xmark & \xmark & \xmark & \xmark & \cmark \\
    \bottomrule
    \end{tabular}
    
    \caption{Comparison between our \textit{MultiFloodSynth} and related datasets. $^\dagger$Dataset will be available under the acceptance.}
    \label{tab:dataset_comparison}
\end{table*}

\begin{figure}[t]
  \centering
  \begin{subfigure}{0.23\textwidth}
    \includegraphics[width=\linewidth]{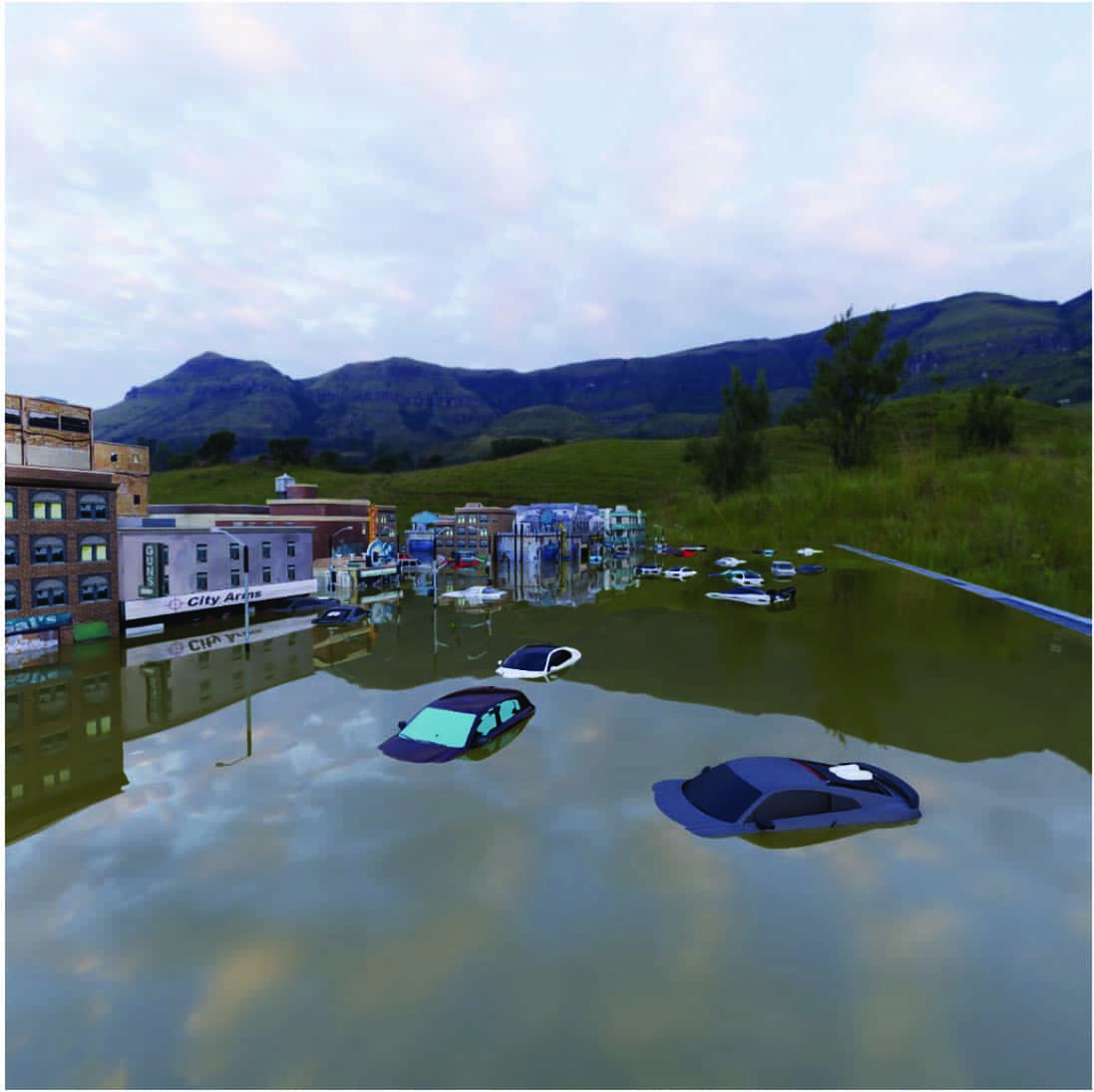}
    \caption{Sample of \textit{flooded} case.}
    \label{fig:flood_sample}
  \end{subfigure} 
  \begin{subfigure}{0.23\textwidth}
    \includegraphics[width=\linewidth]{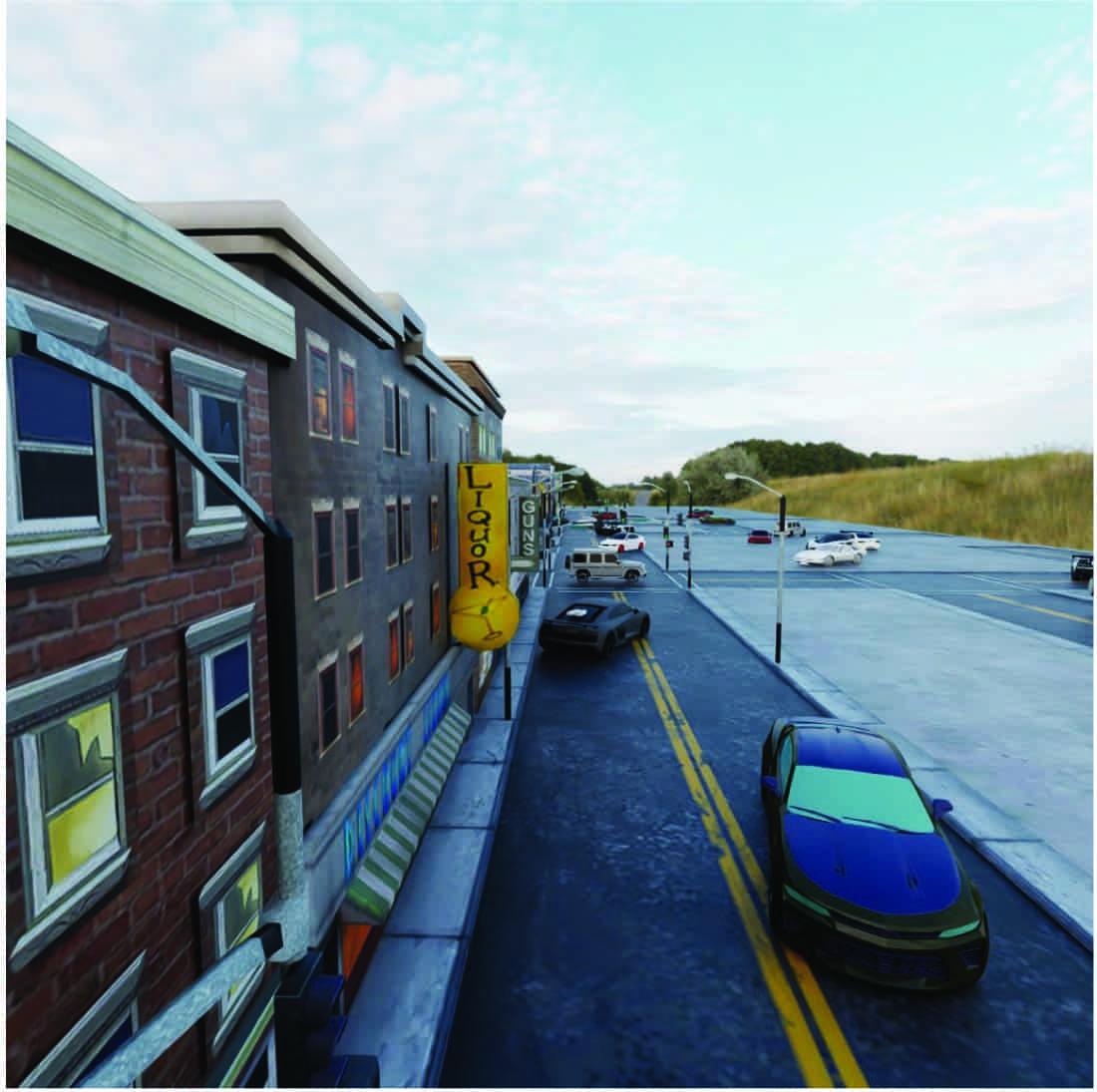}
    \caption{Sample of \textit{non-flooded} case.}
    \label{fig:non_flood_sample}
  \end{subfigure}
  
  \caption{Sample of our \textit{MultiFloodSynth}.}
  \label{fig:sampleofflooddataset}
\end{figure}

\subsection{Depicting Flood Scenarios in Virtual Simulator}

In contrast to existing synthetic generation works, our framework is capable of controlling some parameters to composite final virtual scene as discussed in former subsection. It allows the user to control the scene for user-wanted structure. Based on our exploration with several real-world data \cite{floodzhong2024detection, floodgao2024measuring, floodwu2024identification}, we observed that following settings play a crucial role to determine the virtual scene: \textit{urban setting}, \textit{flood settings} which heavily affects to semantic feature in neural network. Detailed parameters are listed in Table \ref{tab:parameters}.

Furthermore, since it is quite cumbersome to search several 3d objects for scene composition, we adopt image-to-3d model \cite{refwu2024unique3d} to generate 3D objects by inputting web-crawled car image. To avoid quality degradation and blurry texture, image-to-3d is used than text-to-3d \cite{lin2023magic3d}. In the case of layout and building, we utilize 3D city generation \cite{refxie2024citydreamer} results for base layout.

\thispagestyle{plain}

\subsection{Simulating Flood Wave}

In flood hazard situation, flood simulating is pretty important factor which determine the flood level   \cite{wan2024automatic} and annotation-level. Some attributes (\textit{e.g.}, reflection, roughness, opacity, specular, texture) of flood object will directly affect to extract training feature by neural network. In this regards, we also simulate flood dynamics and visual appearance as shown in Fig. \ref{fig:flood_simulation}. Three factors, \textit{main wave, wave foam, gravity}, decide the visual magnitude of level-of-wave as dynamics. Other factors (\textit{e.g.}, wave size, depth, height) change the appearance of flood as static component.

\subsection{\textit{MultiFloodSynth} Generation}

\begin{figure}[t]
  \centering
  \begin{subfigure}{0.15\textwidth}
    \includegraphics[width=\linewidth]{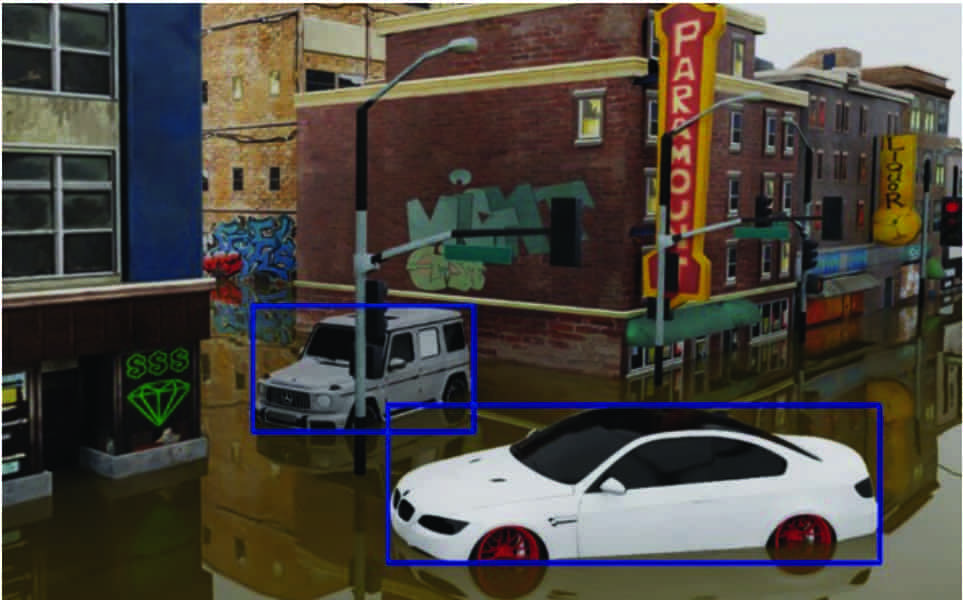}
    \caption{2D bounding box.}
    \label{fig:flood_sample}
  \end{subfigure}
  \begin{subfigure}{0.15\textwidth}
    \includegraphics[width=\linewidth]{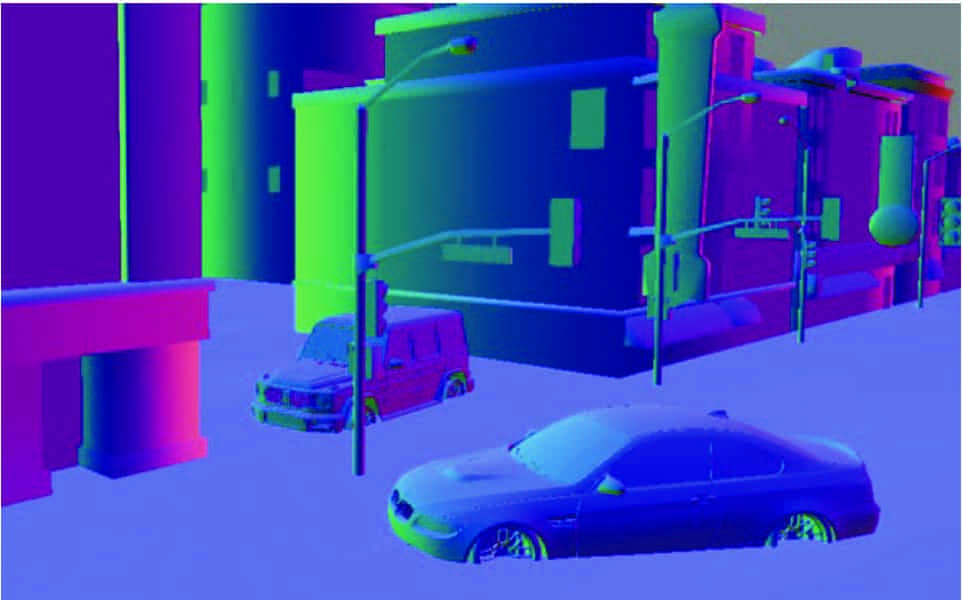}
    \caption{Normal map.}
    \label{fig:non_flood_sample}
  \end{subfigure}
  \begin{subfigure}{0.15\textwidth}
    \includegraphics[width=\linewidth]{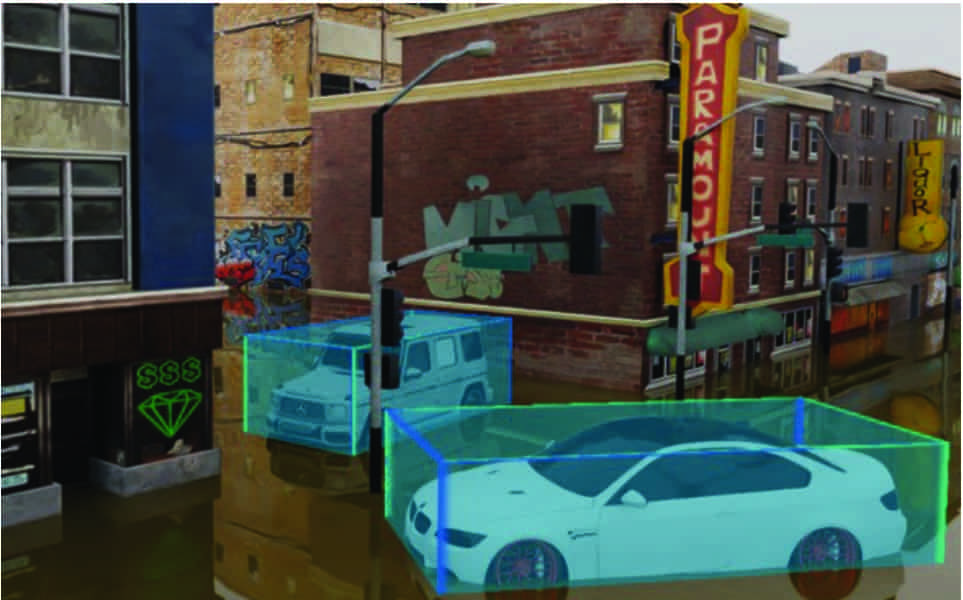}
    \caption{3D bounding box.}
    \label{fig:non_flood_sample}
  \end{subfigure}

  \begin{subfigure}{0.15\textwidth}
    \includegraphics[width=\linewidth]{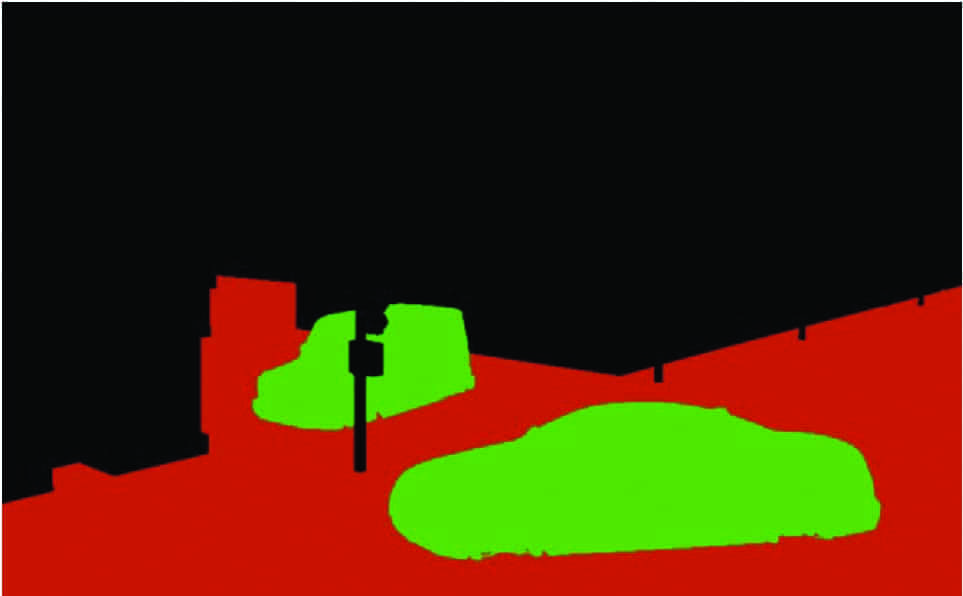}
    \caption{Semantic seg.}
    \label{fig:non_flood_sample}
  \end{subfigure}
  \begin{subfigure}{0.15\textwidth}
    \includegraphics[width=\linewidth]{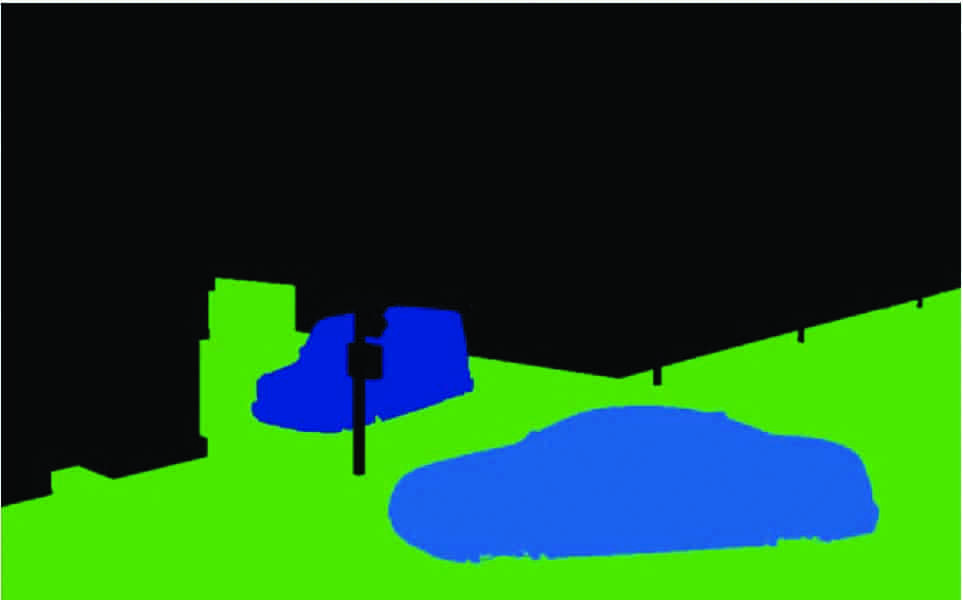}
    \caption{Instance seg.}
    \label{fig:flood_sample}
  \end{subfigure}
  \begin{subfigure}{0.15\textwidth}
    \includegraphics[width=\linewidth]{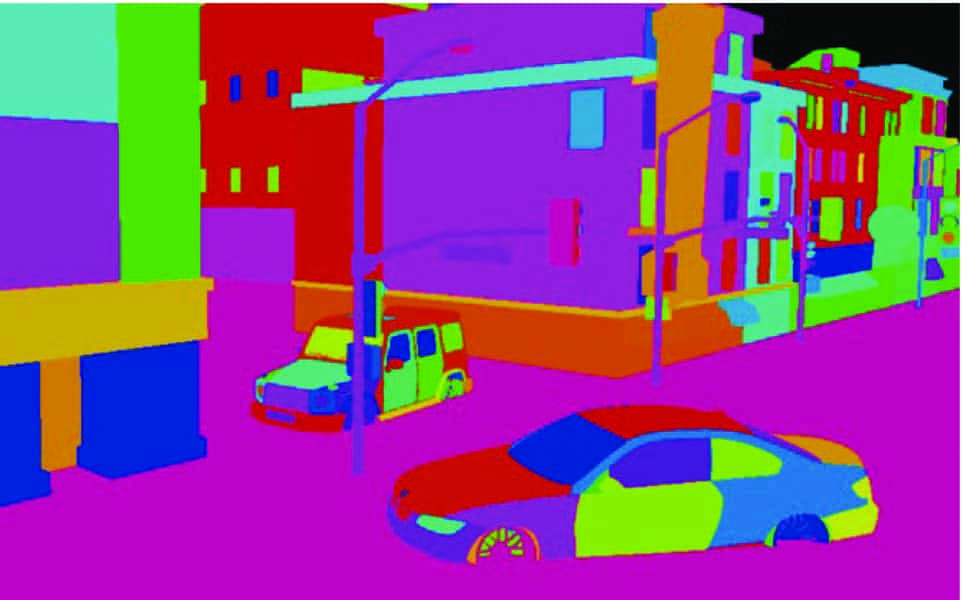}
    \caption{Fine-grained seg.}
    \label{fig:flood_sample}
  \end{subfigure}

   \begin{subfigure}[t]{0.15\textwidth}
    \includegraphics[width=\linewidth]{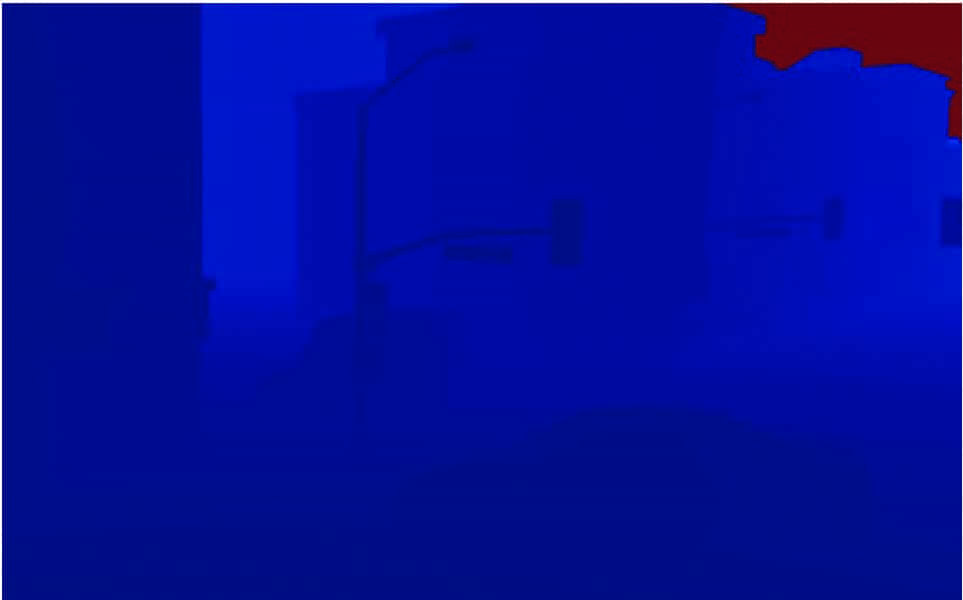}
    \caption{Depth map.}
    \label{fig:flood_sample}
  \end{subfigure}
  \begin{subfigure}[t]{0.15\textwidth}
    \includegraphics[width=\linewidth]{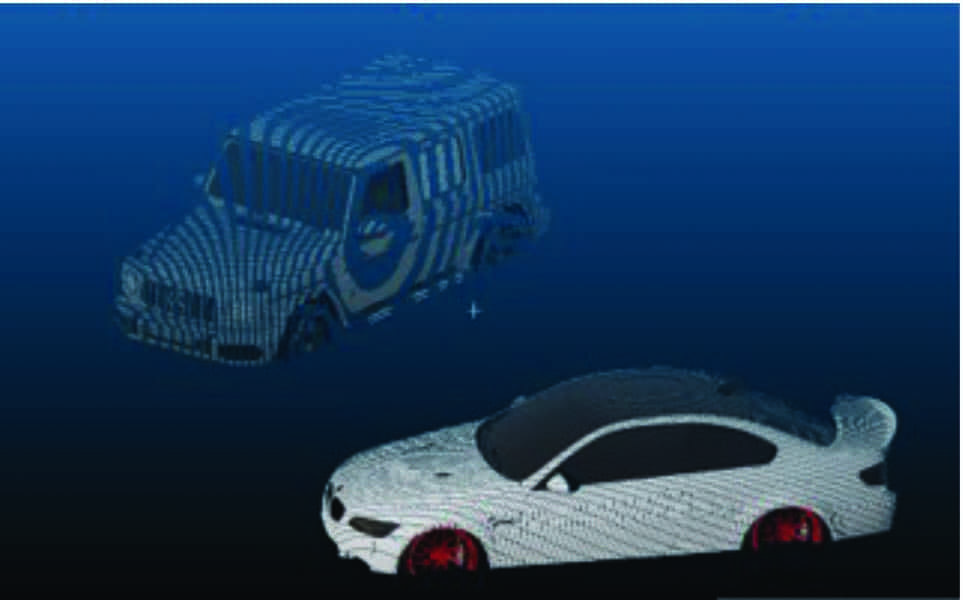}
    \caption{Point cloud.}
    \label{fig:flood_sample}
  \end{subfigure}
  \begin{subfigure}[t]{0.15\textwidth}
    \includegraphics[width=\linewidth]{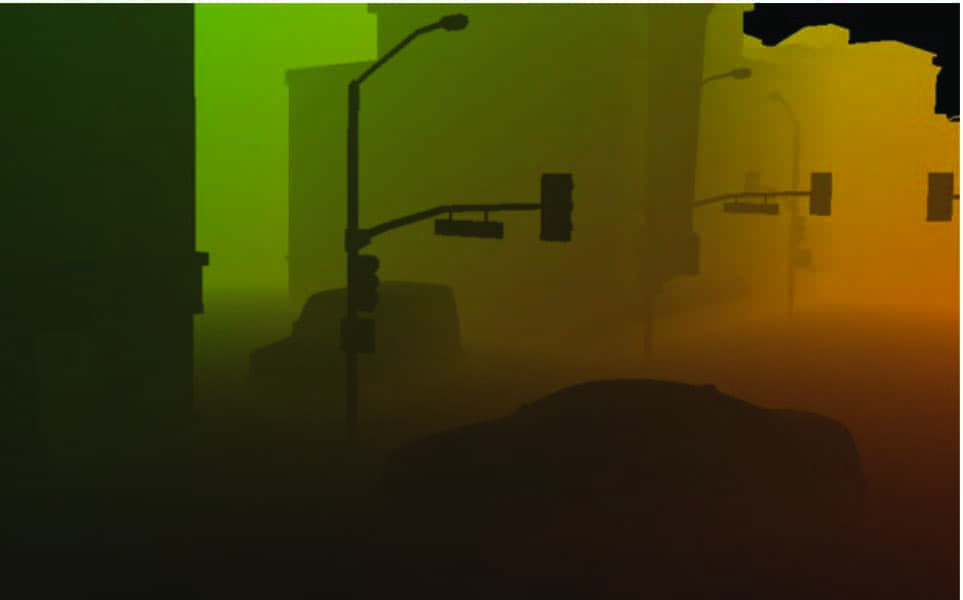}
    \caption{Camera 3D pos.}
    \label{fig:non_flood_sample}
  \end{subfigure}
  
  \caption{Richness of annotation type of our \textit{MultiFloodSynth}.}
  \label{fig:rich_annotations}
\end{figure}

For flood synthetic generation, we construct flood environments based on each objects by varying some parameters (Tab. \ref{tab:parameters}) for diverse data distribution. Flood level is attributed into 5 level as multi-classes. To enhance the quality and domain similarity, we adopt domain randomization \cite{rawal2023synthetic} in all the objects (e.g., light, camera view, object position, etc). Each object is randomly located in every generation pipeline to avoid bias and sparsity of dataset and to include some crucial corner cases. As a result, our \textit{MultiFloodSynth} consists of a total of 70,117 images, with 14,593 \textit{non-flooded} images and 55,524 \textit{flooded} images. Samples are illustrated in Fig. \ref{fig:sampleofflooddataset}. Total image and instance for each class is listed in Tab. \ref{tab:statistics_synth_data}. For multi-type of annotations, we extract 9 types (\textit{i.e.}, semantic/instance/fine-grained segmentation, 2D/3D bounding box) of paired synthetic scene as shown in Fig. \ref{fig:rich_annotations}. For segmentation map, we allocate the label into car and flood. Differences between competing flood datasets are listed in Tab. \ref{tab:dataset_comparison}.

\section{Experiments}

\subsection{Environmental Settings}
For detection model, we choose YOLOv10\cite{wang2024yolov10}. For training hyperparameters, we set batch size as 256, learning rate as 0.001 with 100 epochs. Image-to-3D model is used with Unique3D \cite{refwu2024unique3d}. Our virtual environment is based on NVIDIA Omniverse simulator.

\thispagestyle{plain}

\begin{table}[] 
    \centering
    \begin{tabular}{c | c | c | c}
    \toprule
    \multicolumn{2}{c|}{\textbf{Class}} & \textbf{\# of Images} & \textbf{Instance} \\
    \midrule
    \textit{Non-flooded} & Level 0 & 14,593 & 37,662 \\
    \midrule
    \multirow{4}{*}{\textit{Flooded}} & Level 1 & 17,485 & 55,624 \\
    & Level 2 & 14,541 & 36,141 \\
    & Level 3 & 12,837 & 61,132 \\
    & Level 4 & 10,661 & 24,476 \\
    \midrule
    \multicolumn{2}{c|}{Total} & 70,117 & 215,035 \\
    \bottomrule
    \end{tabular}
    
    \caption{Summary of our \textit{MultiFloodSynth} composition.}
    \label{tab:statistics_synth_data}
\end{table}

\begin{table}[t]
    
    \begin{subtable}[t]{1\linewidth}
    \centering
    \begin{tabular}{ c | c  c  c  c}
    \toprule
    Training & PR(\%)$\uparrow$ & RC(\%)$\uparrow$ & $\text{mAP}_\text{50}\uparrow$ & $\text{mAP}_\text{50-95}\uparrow$ \\
    \midrule
    $\mathcal{D}_\text{real}$ & 37.61 & \textbf{43.81}& 33.21 & 22.92 \\
    \midrule
    $\mathcal{D}_\text{synth}$ & 19.36 & 7.21 & 7.41 & 4.32 \\
    \midrule
     \cellcolor{cyan!20} & \cellcolor{cyan!20} & \cellcolor{cyan!20} & \cellcolor{cyan!20} & \cellcolor{cyan!20}\\
     \multirow{-2}{*}{\shortstack{$\mathcal{D}_\text{real}$ $+$ \\ $\mathcal{D}_\text{synth}$}}
     \cellcolor{cyan!20} & \cellcolor{cyan!20}\multirow{-2}{*}{\textbf{40.97}}  & \cellcolor{cyan!20}\multirow{-2}{*}{41.75}  & \cellcolor{cyan!20}\multirow{-2}{*}{\textbf{35.78}}  & \cellcolor{cyan!20}\multirow{-2}{*}{\textbf{23.16}} \\
    \bottomrule
    \end{tabular}
    \caption{Results on YOLOv10-N.}
    \label{model_n}
    \end{subtable}

    \hspace{\fill}

    \begin{subtable}[t]{1\linewidth}
    \centering
    \begin{tabular}{ c | c  c  c  c}
    \toprule
    Training & PR(\%)$\uparrow$ & RC(\%)$\uparrow$ & $\text{mAP}_\text{50}\uparrow$ & $\text{mAP}_\text{50-95}\uparrow$ \\
    \midrule
    $\mathcal{D}_\text{real}$ & 58.92 & \textbf{58.12}  & 56.66 & 40.64 \\
    \midrule
    $\mathcal{D}_\text{synth}$ & 15.17  & 14.41  & 6.94 & 3.97 \\
    \midrule
     \cellcolor{cyan!20} & \cellcolor{cyan!20} & \cellcolor{cyan!20} & \cellcolor{cyan!20} & \cellcolor{cyan!20}\\
     \multirow{-2}{*}{\shortstack{$\mathcal{D}_\text{real}$ $+$ \\ $\mathcal{D}_\text{synth}$}}
     \cellcolor{cyan!20} & \cellcolor{cyan!20}\multirow{-2}{*}{\textbf{61.16}}  & \cellcolor{cyan!20}\multirow{-2}{*}{57.77}  & \cellcolor{cyan!20}\multirow{-2}{*}{\textbf{58.61}}  & \cellcolor{cyan!20}\multirow{-2}{*}{\textbf{42.71}} \\
    \bottomrule
    \end{tabular}
    \caption{Results on YOLOv10-B.}
    \label{model_n}
    \end{subtable}

    \caption{Classification performance according to the training dataset. PR, RC, mAP denote precision, recall, mean average precision, respectively. Best score is denoted as \textbf{bold-font}.}
    \label{tab:comparison_detection}
\end{table}

\subsection{Comparison on Detection Performance}

To evaluate the superiority of our \textit{MultiFloodSynth}, we compare the flood detection performance by varying the training dataset. Based on \cite{wan2024automatic}, we denotes previous real dataset as $\mathcal{D}_\text{real}$ and our dataset as $\mathcal{D}_\text{synth}$. Then, we train the model with different training configuration as: (1) $\mathcal{D}_\text{real}$, (2), $\mathcal{D}_\text{synth}$, (3) $\mathcal{D}_\text{real} + \mathcal{D}_\text{synth}$. For evaluation metric, we include precision, recall, mAP at 50 and 50-95. In evaluation, we train two size models, YOLOv10-N (\textit{small size}) and YOLOv10-B (\textit{large size}).

The results are shown in Tab. \ref{tab:comparison_detection}. It show that the real-world data training outperformed the synthetic data training. However, training mixing two dataset ($\mathcal{D}_\text{real} + \mathcal{D}_\text{synth}$) demonstrated improved performance. To conclude, our \textit{MultiFloodSynth} boost the detection performance in flood hazard recognition task with consistent annotation while also alleviating the cost of burden data collection process.

\subsection{Evaluation of \textit{MultiFloodSynth}}

Furthermore, we intend to evaluate our \textit{MultiFloodSynth} in the perspective of realism compared with real dataset. To do that, we borrow the recent synthetic dataset evaluation metric, \textit{Realistic Score} which is proposed in urban world generation in \cite{shang2024urbanworld}. By randomly selecting 1K samples in each dataset, we average the score. We normalized the scores of $\mathcal{D}_\text{synth}$ based on the scores of the real dataset. As shown in Tab. \ref{tab:realistic_score}, our dataset showed $93.17\%$ plausibility which is similar level of realism compared with real dataset.

\begin{table}[t]
    \centering
    \begin{tabular}{ l | c }
    \toprule
    \textbf{Data} & \textbf{Realistic Score}(\%)$\uparrow$ \\
    \midrule
    $\mathcal{D}_\text{real}$      & 100{ \scriptsize(7.18)} \\
    $\mathcal{D}_\text{synth}$ & 93.17{ \scriptsize(6.69)} \\
    \bottomrule
    \end{tabular}

    \caption{Realism of synthetic data compared with real data.}
    \label{tab:realistic_score}
\end{table}

\begin{figure}[t]
    \centerline{\includegraphics[width=0.8\linewidth]{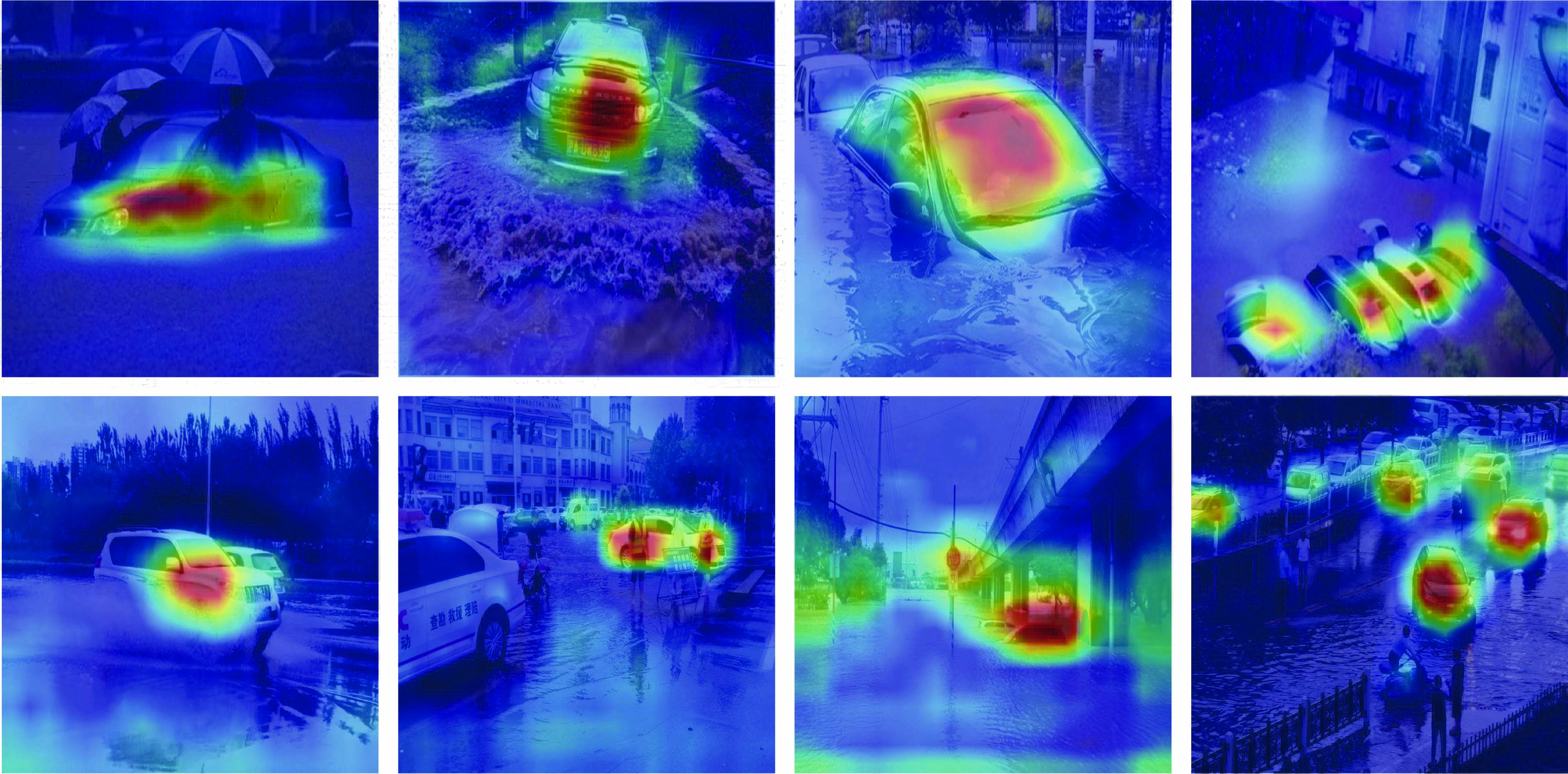}}
    \caption{Visualization of EigenCAM for explainability. A red part indicates that the model considers that part to be important evidence for decision.}
    \label{fig:eigencam}
\end{figure}

\subsection{Explanation of Flood Detection Model}

For explainability of detection, we adopt recent XAI method, 
EigenCAM\cite{muhammad2020eigen} which can analyze some specific parts of an input image that play a crucial role in the model decision. As shown in Fig. \ref{fig:eigencam}, our \textit{MultiFloodSynth} played a crucial role in training and thus enables model to recognize important evidence from images.

\section{Conclusions}
In this paper, we have presented synthetic dataset generation pipeline for urban flood detection into 5 levels and evaluated the utility of our \textit{MultiFloodSynth}. For faithful synthetic dataset, we charactersize several parameters including not only environmental settings, \textit{e.g.}, lighting color, camera position, but also flood simulation factors, \textit{e.g.}, roughness, texture, opacity, etc. To mitigate the data bias and domain gap, we adopt domain randomization and vary the above parameters in each generation pipeline. Moreover, for the sake of efficiency of pipeline, we leverage the recent generation techniques, image-to-3D generation and urban city generation, for 3D object and base layout of virtual flood world, respectively. Experimental results demonstrated that the model trained with our synthetic dataset and real-world dataset show enhanced detection performance in object-localization based flood-level recognition. In addition, our \textit{MultiFloodSynth} showed on-par realism compared with real dataset in terms of \textit{Realistic Score} metric. To conclude, our \textit{MultiFloodSynth} generated from our parameter-controllable flood environment can serve as a valuable training dataset, alternating data requirements of real-world dataset. 

\thispagestyle{plain}

\section{Acknowledgments}
This research was supported by Culture, Sports and Tourism R\&D Program through the Korea Creative Content Agency grant funded by Ministry of Culture, Sports and  Tourism in 2024 (Project Name : Developing Professionals for R\&D in Contents Production Based on Generative AI and Cloud, Project Number : RS-2024-00352578, Contribution Rate: 100\%) and Artificial intelligence industrial convergence cluster development project funded by the Ministry of Science and ICT(MSIT, Korea) \& Gwangju Metropolitan City.

\thispagestyle{plain}

\bibliography{aaai25}

\begin{thebibliography}{36}
\providecommand{\natexlab}[1]{#1}

\bibitem[{Amador~Herrera et~al.(2024)Amador~Herrera, Klein, Liu, Pa{\l}ubicki, Pirk, and Michels}]{amador2024cyclogenesis}
Amador~Herrera, J.~A.; Klein, J.; Liu, D.; Pa{\l}ubicki, W.; Pirk, S.; and Michels, D.~L. 2024.
\newblock Cyclogenesis: Simulating Hurricanes and Tornadoes.
\newblock \emph{ACM Transactions on Graphics (TOG)}, 43(4): 1--16.

\bibitem[{Chaudhary et~al.(2020)Chaudhary, D’Aronco, Leit{\~a}o, Schindler, and Wegner}]{chaudhary2020water}
Chaudhary, P.; D’Aronco, S.; Leit{\~a}o, J.~P.; Schindler, K.; and Wegner, J.~D. 2020.
\newblock Water level prediction from social media images with a multi-task ranking approach.
\newblock \emph{ISPRS Journal of Photogrammetry and Remote Sensing}, 167: 252--262.

\bibitem[{Delussu, Putzu, and Fumera(2024)}]{delussu2024synthetic}
Delussu, R.; Putzu, L.; and Fumera, G. 2024.
\newblock Synthetic Data for Video Surveillance Applications of Computer Vision: A Review.
\newblock \emph{International Journal of Computer Vision}, 1--37.

\bibitem[{Ebadi et~al.(2022)Ebadi, Dhakad, Vishwakarma, Wang, Jhang, Chociej, Crespi, Thaman, and Ganguly}]{ebadi2022psp}
Ebadi, S.~E.; Dhakad, S.; Vishwakarma, S.; Wang, C.; Jhang, Y.-C.; Chociej, M.; Crespi, A.; Thaman, A.; and Ganguly, S. 2022.
\newblock PSP-HDRI $+ $: A Synthetic Dataset Generator for Pre-Training of Human-Centric Computer Vision Models.
\newblock arXiv:2207.05025.

\bibitem[{Gao et~al.(2024)Gao, Yang, Gao, Shao, Wei, and Xu}]{floodgao2024measuring}
Gao, K.; Yang, Z.; Gao, X.; Shao, W.; Wei, H.; and Xu, T. 2024.
\newblock Measuring urban waterlogging depths from video images based on reference objects.
\newblock \emph{Journal of Flood Risk Management}, 17(1): e12948.

\bibitem[{Greff et~al.(2022)Greff, Belletti, Beyer, Doersch, Du, Duckworth, Fleet, Gnanapragasam, Golemo, Herrmann, Kipf, Kundu, Lagun, Laradji, Liu, Meyer, Miao, Nowrouzezahrai, Oztireli, Pot, Radwan, Rebain, Sabour, Sajjadi, Sela, Sitzmann, Stone, Sun, Vora, Wang, Wu, Yi, Zhong, and Tagliasacchi}]{ref25Greff2022KubricAS}
Greff, K.; Belletti, F.; Beyer, L.; Doersch, C.; Du, Y.; Duckworth, D.; Fleet, D.~J.; Gnanapragasam, D.; Golemo, F.; Herrmann, C.; Kipf, T.; Kundu, A.; Lagun, D.; Laradji, I.~H.; Liu, H.-T.; Meyer, H.; Miao, Y.; Nowrouzezahrai, D.; Oztireli, C.; Pot, E.; Radwan, N.; Rebain, D.; Sabour, S.; Sajjadi, M. S.~M.; Sela, M.; Sitzmann, V.; Stone, A.; Sun, D.; Vora, S.; Wang, Z.; Wu, T.; Yi, K.~M.; Zhong, F.; and Tagliasacchi, A. 2022.
\newblock Kubric: A scalable dataset generator.
\newblock \emph{2022 IEEE/CVF Conference on Computer Vision and Pattern Recognition (CVPR)}, 3739--3751.

\bibitem[{Hamza et~al.(2024)Hamza, Lojo, N{\'u}{\~n}ez-Marcos, and Atutxa}]{hamza2024ali}
Hamza, A.; Lojo, A.; N{\'u}{\~n}ez-Marcos, A.; and Atutxa, A. 2024.
\newblock Ali-AUG: Innovative Approaches to Labeled Data Augmentation using One-Step Diffusion Model.
\newblock arXiv:2410.18678.

\bibitem[{Hao et~al.(2024)Hao, Han, Jiang, Li, Wu, Zhong, Zhou, and Tang}]{hao2024synthetic}
Hao, S.; Han, W.; Jiang, T.; Li, Y.; Wu, H.; Zhong, C.; Zhou, Z.; and Tang, H. 2024.
\newblock Synthetic data in AI: Challenges, applications, and ethical implications.
\newblock arXiv:2401.01629.

\bibitem[{Hong et~al.(2024)Hong, Hamdan, Zhao, Ye, Pan, and Cetin}]{hong2024wildfire}
Hong, Z.; Hamdan, E.; Zhao, Y.; Ye, T.; Pan, H.; and Cetin, A.~E. 2024.
\newblock Wildfire detection via transfer learning: a survey.
\newblock \emph{Signal, Image and Video Processing}, 18(1): 207--214.

\bibitem[{Hummel and van Kooten(2019)}]{refhummel2019leveraging}
Hummel, M.; and van Kooten, K. 2019.
\newblock Leveraging nvidia omniverse for in situ visualization.
\newblock In \emph{High Performance Computing: ISC High Performance 2019 International Workshops, Frankfurt, Germany, June 16-20, 2019, Revised Selected Papers 34}, 634--642. Springer.

\bibitem[{Islam et~al.(2024)Islam, Zaheer, Mahmood, and Nandakumar}]{islam2024diffusemix}
Islam, K.; Zaheer, M.~Z.; Mahmood, A.; and Nandakumar, K. 2024.
\newblock DiffuseMix: Label-Preserving Data Augmentation with Diffusion Models.
\newblock In \emph{Proceedings of the IEEE/CVF Conference on Computer Vision and Pattern Recognition}, 27621--27630.

\bibitem[{Jung et~al.(2024)Jung, Byun, Kim, Amin, and Seo}]{refjung2024harnessing}
Jung, Y.; Byun, S.; Kim, B.; Amin, S.~U.; and Seo, S. 2024.
\newblock Harnessing synthetic data for enhanced detection of Pine Wilt Disease: An image classification approach.
\newblock \emph{Computers and Electronics in Agriculture}, 218: 108690.

\bibitem[{Karanjit, Pally, and Samadi(2023)}]{floodkaranjit2023floodimg}
Karanjit, R.; Pally, R.; and Samadi, S. 2023.
\newblock FloodIMG: flood image DataBase system.
\newblock \emph{Data in brief}, 48: 109164.

\bibitem[{Khullar et~al.(2023)Khullar, Sokhandan, Kulkarni, and Shah}]{khullar2023synthetic}
Khullar, D.; Sokhandan, N.; Kulkarni, N.; and Shah, Y. 2023.
\newblock Synthetic data generation for scarce road scene detection scenarios.

\bibitem[{Kim et~al.(2024)Kim, Lam, Lee, and Ok}]{kim2024early}
Kim, H.-C.; Lam, H.-K.; Lee, S.-H.; and Ok, S.-Y. 2024.
\newblock Early Fire Detection System by Using Automatic Synthetic Dataset Generation Model Based on Digital Twins.
\newblock \emph{Applied Sciences}, 14(5): 1801.

\bibitem[{Kokosza et~al.(2024)Kokosza, Wrede, Gonzalez~Esparza, Makowski, Liu, Michels, Pirk, and Palubicki}]{kokosza2024scintilla}
Kokosza, A.; Wrede, H.; Gonzalez~Esparza, D.; Makowski, M.; Liu, D.; Michels, D.~L.; Pirk, S.; and Palubicki, W. 2024.
\newblock Scintilla: Simulating Combustible Vegetation for Wildfires.
\newblock \emph{ACM Transactions on Graphics (TOG)}, 43(4): 1--21.

\bibitem[{Lee, Shin, and Lee(2024)}]{lee2024learning}
Lee, K.; Shin, U.; and Lee, B.-U. 2024.
\newblock Learning to Control Camera Exposure via Reinforcement Learning.
\newblock In \emph{Proceedings of the IEEE/CVF Conference on Computer Vision and Pattern Recognition}, 2975--2983.

\bibitem[{Lin et~al.(2023)Lin, Gao, Tang, Takikawa, Zeng, Huang, Kreis, Fidler, Liu, and Lin}]{lin2023magic3d}
Lin, C.-H.; Gao, J.; Tang, L.; Takikawa, T.; Zeng, X.; Huang, X.; Kreis, K.; Fidler, S.; Liu, M.-Y.; and Lin, T.-Y. 2023.
\newblock Magic3d: High-resolution text-to-3d content creation.
\newblock In \emph{Proceedings of the IEEE/CVF Conference on Computer Vision and Pattern Recognition}, 300--309.

\bibitem[{Lo et~al.(2021)Lo, Wu, Chang, Tseng, Lin, and Lin}]{floodlo2021deep}
Lo, S.-W.; Wu, J.-H.; Chang, J.-Y.; Tseng, C.-H.; Lin, M.-W.; and Lin, F.-P. 2021.
\newblock Deep sensing of urban waterlogging.
\newblock \emph{IEEE Access}, 9: 127185--127203.

\bibitem[{Mittal et~al.(2023)Mittal, Yu, Yu, Liu, Rudin, Hoeller, Yuan, Singh, Guo, Mazhar et~al.}]{mittal2023orbit}
Mittal, M.; Yu, C.; Yu, Q.; Liu, J.; Rudin, N.; Hoeller, D.; Yuan, J.~L.; Singh, R.; Guo, Y.; Mazhar, H.; et~al. 2023.
\newblock Orbit: A unified simulation framework for interactive robot learning environments.
\newblock \emph{IEEE Robotics and Automation Letters}, 8(6): 3740--3747.

\bibitem[{Muhammad and Yeasin(2020)}]{muhammad2020eigen}
Muhammad, M.~B.; and Yeasin, M. 2020.
\newblock Eigen-cam: Class activation map using principal components.
\newblock In \emph{2020 international joint conference on neural networks (IJCNN)}, 1--7. IEEE.

\bibitem[{Pally and Samadi(2022)}]{floodpally2022application}
Pally, R.; and Samadi, S. 2022.
\newblock Application of image processing and convolutional neural networks for flood image classification and semantic segmentation.
\newblock \emph{Environmental modelling \& software}, 148: 105285.

\bibitem[{Rawal, Sompura, and Hintze(2023)}]{rawal2023synthetic}
Rawal, P.; Sompura, M.; and Hintze, W. 2023.
\newblock Synthetic data generation for bridging Sim2Real gap in a production environment.
\newblock arXiv:2311.11039.

\bibitem[{Richter, AlHaija, and Koltun(2022)}]{refrichter2022enhancing}
Richter, S.~R.; AlHaija, H.~A.; and Koltun, V. 2022.
\newblock Enhancing photorealism enhancement.
\newblock \emph{IEEE Transactions on Pattern Analysis and Machine Intelligence}, 45(2): 1700--1715.

\bibitem[{Schieber et~al.(2024)Schieber, Demir, Kleinbeck, Yang, and Roth}]{refschieber2024indoor}
Schieber, H.; Demir, K.~C.; Kleinbeck, C.; Yang, S.~H.; and Roth, D. 2024.
\newblock Indoor synthetic data generation: A systematic review.
\newblock \emph{Computer Vision and Image Understanding}, 103907.

\bibitem[{Shang et~al.(2024)Shang, Lin, Zheng, Fan, Ding, Feng, Chen, Tian, and Li}]{shang2024urbanworld}
Shang, Y.; Lin, Y.; Zheng, Y.; Fan, H.; Ding, J.; Feng, J.; Chen, J.; Tian, L.; and Li, Y. 2024.
\newblock UrbanWorld: An Urban World Model for 3D City Generation.
\newblock arXiv:2407.11965.

\bibitem[{Valvano et~al.(2024)Valvano, Agostino, De~Magistris, Graziano, and Veneri}]{valvano2024controllable}
Valvano, G.; Agostino, A.; De~Magistris, G.; Graziano, A.; and Veneri, G. 2024.
\newblock Controllable Image Synthesis of Industrial Data using Stable Diffusion.
\newblock In \emph{Proceedings of the IEEE/CVF Winter Conference on Applications of Computer Vision}, 5354--5363.

\bibitem[{Wan et~al.(2024)Wan, Qin, Shen, Yang, Yan, Zhang, Yang, Xue, and Wang}]{wan2024automatic}
Wan, J.; Qin, Y.; Shen, Y.; Yang, T.; Yan, X.; Zhang, S.; Yang, G.; Xue, F.; and Wang, Q.~J. 2024.
\newblock Automatic detection of urban flood level with YOLOv8 using flooded vehicle dataset.
\newblock \emph{Journal of Hydrology}, 639: 131625.

\bibitem[{Wang et~al.(2024{\natexlab{a}})Wang, Chen, Liu, Chen, Lin, Han, and Ding}]{wang2024yolov10}
Wang, A.; Chen, H.; Liu, L.; Chen, K.; Lin, Z.; Han, J.; and Ding, G. 2024{\natexlab{a}}.
\newblock Yolov10: Real-time end-to-end object detection.
\newblock arXiv:2405.14458.

\bibitem[{Wang et~al.(2024{\natexlab{b}})Wang, Draghi, Rotalinti, Lunn, and Myles}]{wang2024high}
Wang, Z.; Draghi, B.; Rotalinti, Y.; Lunn, D.; and Myles, P. 2024{\natexlab{b}}.
\newblock High-fidelity synthetic data applications for data augmentation.

\bibitem[{Wu et~al.(2024{\natexlab{a}})Wu, Liu, Cai, Yan, Wang, Hu, Duan, and Ma}]{refwu2024unique3d}
Wu, K.; Liu, F.; Cai, Z.; Yan, R.; Wang, H.; Hu, Y.; Duan, Y.; and Ma, K. 2024{\natexlab{a}}.
\newblock Unique3D: High-Quality and Efficient 3D Mesh Generation from a Single Image.
\newblock arXiv:2405.20343.

\bibitem[{Wu et~al.(2024{\natexlab{b}})Wu, Liu, Zhang, Zhang, Wang, Tong, Li, and Zhang}]{floodwu2024identification}
Wu, L.; Liu, Y.; Zhang, J.; Zhang, B.; Wang, Z.; Tong, J.; Li, M.; and Zhang, A. 2024{\natexlab{b}}.
\newblock Identification of flood depth levels in urban waterlogging disaster caused by rainstorm using a CBAM-improved ResNet50.
\newblock \emph{Expert Systems with Applications}, 124382.

\bibitem[{Xie et~al.(2024)Xie, Chen, Hong, and Liu}]{refxie2024citydreamer}
Xie, H.; Chen, Z.; Hong, F.; and Liu, Z. 2024.
\newblock Citydreamer: Compositional generative model of unbounded 3d cities.
\newblock In \emph{Proceedings of the IEEE/CVF Conference on Computer Vision and Pattern Recognition}, 9666--9675.

\bibitem[{Zhang et~al.(2024)Zhang, Zhou, Wang, Luo, Wang, Li, Yin, Zhang, and Peng}]{refzhang2024cityx}
Zhang, S.; Zhou, M.; Wang, Y.; Luo, C.; Wang, R.; Li, Y.; Yin, X.; Zhang, Z.; and Peng, J. 2024.
\newblock CityX: Controllable Procedural Content Generation for Unbounded 3D Cities.
\newblock arXiv:2407.17572.

\bibitem[{Zhong et~al.(2024)Zhong, Liu, Zheng, and Zhao}]{floodzhong2024detection}
Zhong, P.; Liu, Y.; Zheng, H.; and Zhao, J. 2024.
\newblock Detection of urban flood inundation from traffic images using deep learning methods.
\newblock \emph{Water Resources Management}, 38(1): 287--301.

\bibitem[{Zhu et~al.(2024)Zhu, Li, Liu, Huang, Shan, Ma, and Yuan}]{zhu2024odgen}
Zhu, J.; Li, S.; Liu, Y.; Huang, P.; Shan, J.; Ma, H.; and Yuan, J. 2024.
\newblock ODGEN: Domain-specific Object Detection Data Generation with Diffusion Models.
\newblock arXiv:2405.15199.

\end{thebibliography}

\thispagestyle{plain}

\end{document}